\documentstyle[colacl]{article}

\title{A Simple Approach to Building Ensembles of Naive Bayesian
Classifiers for Word Sense Disambiguation} 

\author{Ted Pedersen\\
Department of Computer Science\\
University of Minnesota Duluth\\
Duluth, MN 55812 USA\\
{\tt tpederse@d.umn.edu}}

\begin{document}
\maketitle

\begin{abstract}
This paper presents a corpus-based approach to word sense 
disambiguation that builds an ensemble of Naive Bayesian classifiers,  
each of which is based on lexical features that represent
co--occurring words in varying sized windows of context. Despite the
simplicity of this approach, empirical results disambiguating the
widely studied nouns  {\it line} and   {\it interest} show that such
an ensemble achieves accuracy rivaling the best previously published
results.     
\end{abstract}

\section{Introduction}

Word sense disambiguation is often cast as a problem in supervised  
learning, where a disambiguator is induced from a corpus of manually  
sense--tagged text using methods from statistics or machine learning. 
These approaches typically represent the context in which each sense--tagged   
instance of a word occurs with a set of linguistically motivated features. 
A learning algorithm induces a representative model from these
features which is employed as a classifier to perform disambiguation. 

This paper presents a corpus--based approach that results in high   
accuracy by combining a number of very simple classifiers 
into an ensemble that performs disambiguation via a majority vote.  This 
is motivated by the observation that enhancing the 
feature set or learning algorithm used in a corpus--based approach 
does not usually improve disambiguation accuracy beyond  what can be
attained with  shallow lexical features and a simple supervised
learning algorithm.   

For example, a Naive  Bayesian classifier \cite{DudaH73} is
based on a blanket  assumption about the interactions among 
features in a sense-tagged corpus and does not learn a representative
model. Despite making such an assumption, this proves 
to be among  the most accurate techniques in  comparative  studies of
corpus--based  word sense  disambiguation methodologies  (e.g.,
\cite{LeacockTV93},  \cite{Mooney96}, \cite{NgL96},
\cite{PedersenB97A}).  These studies  represent the context in which
an ambiguous word occurs with a wide variety  of features. However, when 
the contribution of each type of  feature to  overall accuracy is analyzed 
(eg.  \cite{NgL96}), shallow  lexical features such as co--occurrences and  
collocations prove to be  stronger contributors to accuracy than do 
deeper, linguistically motivated features such as part--of--speech and
verb--object relationships. 

It has also been shown that the combined 
accuracy of an  ensemble  of  multiple  classifiers is often significantly 
greater than  that of any  of  the  individual classifiers  that make up 
the ensemble  (e.g., \cite{Dietterich97}).  In natural language
processing, ensemble techniques have been successfully applied to
part--of--speech tagging (e.g., \cite{BrillH98}) and parsing (e.g.,
\cite{HendersonB99}). When combined with a history
of disambiguation success using shallow lexical  features and
Naive Bayesian classifiers, these findings suggest that word sense
disambiguation might best be improved by combining the output of a
number of such classifiers into an ensemble.  

This paper begins with an introduction to the Naive Bayesian classifier. 
The features used to represent the context in which ambiguous words occur 
are presented, followed by the method for selecting the classifiers to 
include in the ensemble.  Then, the {\it line} and  {\it interest} data is 
described.  Experimental results disambiguating   these words with an
ensemble of Naive Bayesian classifiers are shown to rival previously 
published results. This paper closes with a discussion of the choices made
in formulating this methodology and plans for future work. 

\section{Naive Bayesian Classifiers}

A Naive Bayesian classifier assumes that all the feature variables  
representing a problem are conditionally independent given the value of a 
classification variable.  In  word sense  disambiguation, the context in 
which an ambiguous word occurs is  represented by the feature variables 
$(F_1, F_2, \ldots,F_n)$ and the sense of the ambiguous word is  
represented by the classification variable $(S)$. In this paper, all 
feature variables $F_i$ are  binary and represent whether or not a 
particular word occurs within some  number of words to the left or right 
of an ambiguous word, i.e., a window 
of context. For a Naive Bayesian classifier, the joint probability of   
observing a certain combination of contextual features with a particular
sense is expressed as:
\begin{eqnarray*}
p(F_1, F_2, \ldots, F_{n},S) = p(S) \prod_{i=1}^{n} p(F_i|S)
\end{eqnarray*}

The parameters of this model are $p(S)$ and $p(F_i|S)$. The sufficient
statistics, i.e., the summaries of the data needed for parameter
estimation, are the frequency counts of the events described by the
interdependent variables $(F_i,S)$.  In this paper, these counts 
are the number of sentences in the sense--tagged text where the word  
represented by $F_i$ occurs within some specified window of context of
the ambiguous word when it is used  in sense $S$. 

Any parameter that has a value of zero indicates that the associated word
never occurs with the specified sense value. These zero values are
smoothed by assigning them a very small default probability. Once all the 
parameters  have been estimated, the model has been trained and can be 
used as a classifier to perform disambiguation by determining the most  
probable sense for an ambiguous word, given the context in which it 
occurs. 

\subsection{Representation of Context}

The contextual features used in this paper are binary and indicate if
a given word occurs within some number of words to the left or right of   
the ambiguous word. No additional positional information  is contained
in these features; they simply indicate if the word occurs  within
some number of surrounding  words.

Punctuation and capitalization are removed from the windows of
context. All other lexical items are included in their original form;
no stemming is performed and non--content words remain. 

This representation of context is a  variation on the {\it
bag--of--words} feature set, where  a single window  of context
includes words that occur  to both the left and  right of the
ambiguous word. An early use of this representation is described in 
\cite{GaleCY92}, where word sense disambiguation is performed with 
a Naive Bayesian classifier. The work in this paper differs in that
there are two windows of  context, one representing words that occur
to the  left of the ambiguous  word and another for those to the
right.

\subsection{Ensembles of Naive Bayesian Classifiers}

The left and right windows of context have nine different sizes;  0,
1, 2, 3, 4, 5, 10, 25, and 50 words.  The first step in the ensemble
approach is to train a separate Naive Bayesian classifier for each of
the 81 possible combination of left and right window sizes. 

Naive\_Bayes (l,r) represents a classifier where the model
parameters have been estimated based on frequency counts of shallow 
lexical features from two windows of context; one including $l$ words to 
the left of the ambiguous word and the other including $r$ words to the
right. 
Note that Naive\_Bayes (0,0) includes no words to the left or
right; this classifier acts as a majority classifier that  assigns 
every instance of an ambiguous word to the  most frequent sense in the 
training data. Once the individual classifiers are trained they are 
evaluated using previously held--out test data.  

The crucial step in building an ensemble is selecting the classifiers to
include as members. The approach here is to group the 81 Naive
Bayesian classifiers into general categories representing
the sizes of the windows of  context. There are three such ranges;
{\it narrow} corresponds  to windows 0, 1 and 2 words wide, {\it 
medium} to windows 3, 4, and 5 
words wide, and {\it wide} to windows 10,  25, and 50 words wide.
There are nine possible range categories since there are separate
left and right windows. For example, Naive\_Bayes(1,3) belongs to the 
range category (narrow, medium) since it is based on a one word window 
to the left and a three word window to the right.  The most  accurate   
classifier in each of the nine range categories  is selected  for 
inclusion in the ensemble.  Each of the nine member classifiers votes for 
the most probable sense given the particular context represented by that 
classifier;  the ensemble disambiguates by assigning the sense that  
receives a majority of the votes. 

\section{Experimental Data}

The {\it line} data was created by  \cite{LeacockTV93} by tagging every 
occurrence of {\it line} in the ACL/DCI Wall Street Journal corpus and the 
American Printing House for the Blind corpus with one of six possible 
WordNet senses. These senses and their frequency  distribution are shown 
in Table \ref{fig:linesenses}.  This data has since been used in 
studies by \cite{Mooney96}, \cite{TowellV98}, and \cite{LeacockCM98}. In 
that work, as well as in this paper, a subset of the corpus is utilized such 
that each sense is uniformly distributed; this reduces the accuracy of the 
majority classifier to 17\%. The uniform distribution is created by  
randomly sampling 349 sense--tagged examples  from each sense, resulting  
in a training corpus of 2094 sense--tagged sentences. 

\begin{table}
\begin{center}
\begin{tabular}{|lr|} \hline
sense & count \\ \hline
product & 2218  \\
written or spoken text  & 405  \\
telephone connection & 429  \\
formation of people or things; queue & 349  \\
an artificial division; boundary & 376  \\
a thin, flexible object; cord & 371  \\ \hline
total & 4148 \\ \hline
\end{tabular}
\caption{Distribution of senses for {\it line} -- the experiments
in this paper and previous work use a uniformly distributed
subset of this corpus, where each sense occurs 349 times.}
\label{fig:linesenses}
\end{center}
\end{table}

The {\it interest} data was created by \cite{BruceW94B} by tagging all  
occurrences of {\it interest} in the ACL/DCI Wall Street Journal corpus
with senses from the Longman Dictionary of Contemporary  English. 
This data set was subsequently used for word 
sense disambiguation experiments by \cite{NgL96}, \cite{PedersenBW97}, and 
\cite{PedersenB97A}. The previous studies and this paper use the entire 
2,368 sense--tagged sentence corpus in their experiments. The  senses and 
their frequency  distribution are shown in Table 
\ref{fig:interestsenses}. Unlike {\it line}, the sense distribution is 
skewed; the majority sense occurs in 53\% of the  sentences,
while the  smallest minority sense  occurs in less than 1\%. 

\begin{table}
\begin{center}
\begin{tabular}{|lr|} \hline
sense & count \\ \hline
money paid for the use of money & 1252 \\
a share in a company or business & 500 \\
readiness to give attention & 361 \\
advantage, advancement or favor & 178 \\
activity that one gives attention to & 66 \\
causing attention to be given to & 11 \\ \hline
total & 2368 \\ \hline
\end{tabular}
\caption{Distribution of senses for {\it interest} -- the experiments
in this paper and previous work use the entire corpus, where each 
sense occurs the number of times shown above.}
\label{fig:interestsenses}
\end{center}
\end{table}

\section{Experimental Results}

Eighty-one Naive Bayesian classifiers were trained and tested with
the {\it line} and {\it interest} data. Five--fold cross validation was
employed; all of the sense--tagged examples for a word were randomly 
shuffled and  divided into five equal folds. Four folds were used to train
the Naive Bayesian classifier while the remaining fold was randomly
divided into two equal sized test sets. The first, {\tt devtest}, was
used to evaluate the individual classifiers for inclusion in the
ensemble. The second, {\tt test}, was used to evaluate the accuracy of
the ensemble. Thus the training data for each word consists of 80\% of
the available sense--tagged text, while each of the test sets contains
10\%.  

This process is repeated five times  so that each fold serves as the
source of the test data once. The average accuracy of the individual
Naive Bayesian classifiers across  the five folds is reported  in
Tables \ref{fig:lineresults} and \ref{fig:interestresults}. The
standard deviations were between .01 and .025 and are  not  shown
given their relative consistency.  

Each classifier is based upon a distinct representation of  
context since each employs a different combination of right and left
window sizes. The size and range of the left window of context is 
indicated along the horizontal margin in Tables \ref{fig:lineresults}
and \ref{fig:interestresults} while the right window  size and range is 
shown along the  vertical margin. Thus, the boxes that subdivide each 
table correspond to a particular range category. The classifier
that achieves the highest accuracy in each range category is included 
as a member of the ensemble. In case of a tie,  the classifier with
the smallest total window of context  is included in the ensemble.  

\begin{table*}
\begin{center}
\begin{tabular}{cr|ccc||ccc||ccc|} 
\cline{3-11}
& 50 & .63 & .73 & .80 & .82 & .83 & .83 & .83 & .83 & .83 \\
wide & 25 & .63 & .74 & .80 & .82 & {\it .84} & .83 & .83 & .83 & .83 \\
& 10 & .62 & .75 & {\it .81} & .82 & .83 & .83 & .83 & .83 & {\it .84} \\ 
\cline{3-11}
\cline{3-11}
& 5  & .61 & .75 & .80 & .81 & .82 & .82 & .82 & .82 & {\it .83}\\
medium & 4  & .60 & .73 & {\it .80} & .82 & .82 & .82 & .82 & .82 & .82 \\
& 3  & .58 & .73 & .79 & .82 & {\it .83} & .83 & .82 & .81 & .82 \\
\cline{3-11}
\cline{3-11}
& 2  & .53 & .71 & {\it .79} & .81 & {\it.82} & .82 & {\it .81} &.81 & .81 \\
narrow& 1  & .42 & .68 & .78 & .79 & .80 & .79 & .80 & .81 & .81 \\
& 0  & .14 & .58 & .73 & .77 & .79 & .79 & .79 & .79 & .80 \\
\cline{3-11}
\multicolumn{2}{c}{} &
\multicolumn{1}{c}{0} &
\multicolumn{1}{c}{1} &
\multicolumn{1}{c}{2} &
\multicolumn{1}{c}{3} &
\multicolumn{1}{c}{4} &
\multicolumn{1}{c}{5} &
\multicolumn{1}{c}{10} &
\multicolumn{1}{c}{25} &
\multicolumn{1}{c}{50} \\
\multicolumn{2}{c}{} &
\multicolumn{3}{c}{narrow} &
\multicolumn{3}{c}{medium} &
\multicolumn{3}{c}{wide} \\
%%  & 0    &  1   &    2 &    3 &    4 &    5 &   10 &   25 &   50 \\
\end{tabular}
\caption{Accuracy of Naive Bayesian classifiers for {\it line}
evaluated with the {\tt devtest} data. The italicized accuracies are
associated with the classifiers included in the ensemble, which
attained accuracy of 88\% when evaluated with the {\tt test} data.}
\label{fig:lineresults}
\end{center}
\end{table*}

\begin{table*}
\begin{center}
\begin{tabular}{cr|ccc||ccc||ccc|} 
\cline{3-11}
&50 & .74 & .80 & .82 & .83 & .83 & .83 & .82 & .80 & .81 \\
wide &25 & .73 & .80 & .82 & .83 & .83 & .83 & .81 & .80 & .80 \\
&10 & .75 & .82 & {\it .84} & {\it .84} & .84 & .84 & {\it .82} &.81 & .81 \\
\cline{3-11} 
&5  & .73 & .83 & .85 & .86 & .85 & .85 & .83 & .81 & .81\\
medium&4  & .72 & .83 & .85 & .85 & .84 & .84 & .83 & .81 & .80\\
&3  & .70 & .84 & {\it .86} & {\it .86} & .86 & .85 & {\it .83} & .81 &.80 \\ 
\cline{3-11} 
&2  & .66 & .83 & .85 & .86 & .86 & .84 & {\it .83} & .80 &.80 \\
narrow&1  & .63 & .82 & {\it .85} & .85 & {\it .86} & .85 & .82 & .81 &
.80 \\ 
&0  & .53 & .72 & .77 & .78 & .79 & .77 & .77 & .76 & .75 \\
\cline{3-11}
\multicolumn{2}{c}{} &
\multicolumn{1}{c}{0} &
\multicolumn{1}{c}{1} &
\multicolumn{1}{c}{2} &
\multicolumn{1}{c}{3} &
\multicolumn{1}{c}{4} &
\multicolumn{1}{c}{5} &
\multicolumn{1}{c}{10} &
\multicolumn{1}{c}{25} &
\multicolumn{1}{c}{50} \\
\multicolumn{2}{c}{} &
\multicolumn{3}{c}{narrow} &
\multicolumn{3}{c}{medium} &
\multicolumn{3}{c}{wide} \\
%%  & 0    &  1   &    2 &    3 &    4 &    5 &   10 &   25 &   50 \\
\end{tabular}
\caption{Accuracy of Naive Bayesian classifiers for {\it interest}
evaluated with the {\tt devtest} data. The italicized accuracies are
associated with the classifiers included in the ensemble, which
attained accuracy of 89\% when evaluated with the {\tt test} data.}
\label{fig:interestresults}
\end{center}
\end{table*}

The most accurate single classifier for {\it line} is Naive\_Bayes 
(4,25), which attains accuracy of 84\% The accuracy of the ensemble 
created from the most accurate classifier in each of the range categories 
is 88\%. The single most 
accurate  classifier for {\it interest} is Naive\_Bayes(4,1), which  
attains accuracy of 86\% while the ensemble approach reaches 89\%. 
The increase in accuracy achieved by both  ensembles over  the best
individual classifier is  statistically  significant,  as judged  by
McNemar's test with $p=.01$.  

\subsection{Comparison to Previous Results}

These experiments use the same sense--tagged corpora for {\it interest} 
and {\it line} as previous studies. Summaries of previous results 
in Tables \ref{fig:previousinterestresults} 
and  \ref{fig:previouslineresults} show
that the accuracy of the Naive Bayesian ensemble is comparable to that
of any other approach.  However, due to variations in experimental 
methodologies, it can  not be concluded that the differences among the
most accurate  methods are  statistically significant. 
For  example, in this work five-fold cross  validation is employed to
assess  accuracy while \cite{NgL96} train and  test using 100 randomly
sampled sets  of data. Similar differences in  training  and testing
methodology exist  among the other studies.  Still, the results in
this paper are encouraging due to the simplicity of the approach. 

\begin{table*}
\begin{center}
\begin{tabular}{|l|c|r|r|} \hline
\multicolumn{1}{|c|}{} &
\multicolumn{1}{c|}{accuracy} &
\multicolumn{1}{c|}{method} &
\multicolumn{1}{c|}{feature set} \\ \hline \hline
%%  & accuracy & method & features\\ \hline 
Naive Bayesian Ensemble & 89\% & ensemble of 9 & varying left \& right b-o-w\\ \hline
Ng \& Lee, 1996      & 87\% & nearest neighbor & p-o-s, morph, co-occur \\ 
      &  &  & collocates, verb-obj \\ \hline
Bruce \& Wiebe, 1994 & 78\%   & model selection & p-o-s, morph,
co--occur\\
\hline
Pedersen \& Bruce, 1997 & 78\% & decision tree & p-o-s, morph, co--occur\\ 
                        & 74\% & naive bayes & \\ \hline
\end{tabular}
\caption{Comparison to previous results for {\it interest}}
\label{fig:previousinterestresults}
\end{center}
\end{table*}

\begin{table*}
\begin{center}
\begin{tabular}{|l|c|r|r|} \hline
\multicolumn{1}{|c|}{} &
\multicolumn{1}{c|}{accuracy} &
\multicolumn{1}{c|}{method} &
\multicolumn{1}{c|}{feature set} \\ \hline \hline
%%  & accuracy & method & features\\ \hline 
Naive Bayesian Ensemble       & 88\% & ensemble of 9 & varying left \& 
right b-o-w 
\\ \hline
Towell \& Voorhess, 1998    & 87\% & neural net & local \& topical b-o-w, p-o-s \\
\hline  
Leacock, Chodorow, \& Miller, 1998  & 84\% & naive bayes & local \& topical b-o-w, p-o-s\\ \hline
Leacock, Towell, \& Voorhees, 1993 &  76\% & neural net & 2 sentence b-o-w \\
                             & 72\% & content vector & \\
                             & 71\% & naive bayes & \\ \hline
Mooney, 1996                 & 72\% & naive bayes & 2 sentence b-o-w 
\\
                             & 71\% & perceptron & \\ \hline
\end{tabular}
\caption{Comparison to previous results for {\it line}}
\label{fig:previouslineresults}
\end{center}
\end{table*}

\subsubsection{Interest}

The {\it interest} data was first studied by  \cite{BruceW94B}. They 
employ a representation of context that includes the part--of--speech of 
the two words surrounding  {\it interest}, a 
morphological feature indicating whether or not {\it  interest} is 
singular or plural, and the three most statistically significant 
co--occurring words in the sentence with {\it interest}, as  determined by 
a test of independence.  These features are abbreviated as {\it p-o-s}, 
{\it  morph}, and {\it co--occur} in Table
\ref{fig:previousinterestresults}. A decomposable probabilistic model  is
induced from the  sense--tagged  corpora using a backward sequential
search where candidate  models are  evaluated with  the log--likelihood
ratio test. The selected  model was used as a  probabilistic classifier on
a held--out set of test  data and achieved accuracy of 78\%.  

The {\it interest} data was included in a study by \cite{NgL96}, who
represent the context of an ambiguous word with the part--of--speech of 
three words to the left and right of {\it  interest}, a morphological 
feature indicating if {\it interest} is  singular or plural, an unordered 
set of frequently occurring keywords that  surround {\it interest}, local 
collocations that include {\it interest},  and verb--object syntactic 
relationships. These features are  abbreviated {\it p-o-s}, {\it morph}, 
{\it co--occur}, {\it collocates},  and  {\it verb--obj} in Table 
\ref{fig:previousinterestresults}. A nearest--neighbor classifier was
employed and achieved an average accuracy of 87\% over repeated
trials using randomly drawn training and test sets. 

\cite{PedersenBW97} and \cite{PedersenB97A} present studies that 
utilize the original Bruce and Wiebe feature set and include the 
{\it interest} data. The first compares  a range of probabilistic model
selection methodologies and finds that none  outperform the Naive Bayesian
classifier, which attains  accuracy of 74\%.  The second compares a range 
of  machine learning   algorithms and finds that a decision tree learner
(78\%) and a Naive Bayesian classifier  (74\%) are most accurate.  

\subsubsection{Line}

The {\it line} data was first studied by  \cite{LeacockTV93}. 
They evaluate the disambiguation accuracy of a Naive 
Bayesian classifier, a content vector, and a neural network.  The context 
of an ambiguous word is represented by a bag--of--words where  
the window of context is two sentences wide. This feature set is  
abbreviated as {\it 2 sentence b-o-w} in Table 
\ref{fig:previouslineresults}. 
When the Naive Bayesian classifier is
evaluated words are not  stemmed and capitalization remains. 
However, with the content  vector and the neural network words are
stemmed and words from a stop--list are removed.  
They report no 
significant differences in accuracy among the three approaches;  the Naive 
Bayesian classifier  achieved 71\% accuracy,  the content vector  72\%, 
and the neural network 76\%. 

The {\it line} data was studied again by \cite{Mooney96}, where 
seven different machine learning methodologies are compared. All 
learning algorithms represent the context of an ambiguous word using the 
bag--of--words with a two sentence window of context.  In these 
experiments  words from a stop--list are removed, capitalization is
ignored, and words are stemmed. The two most accurate methods in this 
study proved to  be a Naive Bayesian classifier (72\%) and a perceptron 
(71\%). 

The {\it line} data was recently revisited by both \cite{TowellV98} and  
\cite{LeacockCM98}. The former take an ensemble approach where the 
output from two neural networks is combined; one network is based on a 
representation  of local context while the other represents topical   
context. The latter  utilize a Naive  Bayesian classifier. In both cases 
context is represented by a set of  topical and local features. The 
topical features correspond to the open--class words that occur in a two 
sentence window of context. The local features occur within a window of 
context three words to the left and right of the ambiguous word  and 
include co--occurrence features as well as the part--of--speech of words 
in this window. These features are represented as {\it local} \&
{\it topical  b-o-w} and {\it p-o-s} in Table 
\ref{fig:previouslineresults}. 
\cite{TowellV98} report accuracy  of 87\%  while  \cite{LeacockCM98} 
report accuracy of 84\%.  

\section{Discussion}

The word sense disambiguation ensembles in this paper have the following 
characteristics:
\begin{itemize}
\item The members of the ensemble are Naive Bayesian classifiers, 
\item the context in which an ambiguous word occurs is represented by 
co--occurrence features extracted from  varying sized windows of
surrounding words,  
\item member classifiers are selected for the ensembles based on their 
performance relative to others with comparable window sizes, and 
\item a majority vote of the member classifiers determines the outcome of 
the ensemble. 
\end{itemize}
Each point is discussed below. 

\subsection{Naive Bayesian classifiers}

The Naive Bayesian classifier has emerged as a 
consistently strong performer in a wide range of comparative studies 
of machine learning methodologies.  A recent survey of such results, as 
well as possible explanations for its success, is presented in 
\cite{DomingosP97}.  A similar finding has emerged in word sense
disambiguation, where a number of comparative studies have all reported
that no method achieves significantly greater accuracy than the Naive Bayesian
classifier  (e.g., \cite{LeacockTV93}, \cite{Mooney96}, \cite{NgL96}, 
\cite{PedersenB97A}). 

In many ensemble approaches the member classifiers are learned with
different algorithms that are trained with the same data.  For
example, an ensemble could consist of a decision tree, a neural
network, and a nearest neighbor classifier, all of which are learned
from exactly the same set of training data.  This paper takes a
different approach, where the learning algorithm is the same for all
classifiers but the training data is different. 
This is motivated by the belief that there is more to be gained by
varying the representation of context than there is from using many
different learning algorithms on the same data. This is especially
true in this domain since the Naive Bayesian classifier has a history
of success and since there is no generally agreed upon set of features
that have been shown to be optimal for word sense disambiguation. 

\subsection{Co--occurrence features}

Shallow lexical features such as co--occurrences and collocations are
recognized as potent sources of disambiguation information. While many
other contextual features are often employed, it isn't  clear that
they offer substantial advantages. For example, \cite{NgL96} report
that local collocations alone achieve  80\%  accuracy disambiguating
{\it interest}, while their full set of features result in 87\%.
Preliminary experiments for this paper used feature sets that 
included  collocates, co--occurrences, part--of--speech and
grammatical information for surrounding words. However, it was clear
that no combination of features resulted in disambiguation accuracy
significantly higher than that achieved with co--occurrence features. 

\subsection{Selecting ensemble members}

The most accurate classifier from each of nine possible category ranges is
selected as a member of the ensemble. This is based on preliminary  
experiments that showed that member classifiers with similar sized 
windows of context  often result in little or no overall improvement in 
disambiguation accuracy. This was expected since slight differences in 
window sizes lead to roughly equivalent representations of context
and classifiers that have little opportunity for collective
improvement. For  example, an  ensemble was created for {\it
interest} using the nine classifiers  in the range   category (medium,
medium). The  accuracy  of this ensemble  was 84\%,   slightly  less
than the  most  accurate  individual classifiers in  that  range which
achieved accuracy  of 86\%. 

Early experiments also revealed that an ensemble based on a majority vote 
of all 81 classifiers performed rather poorly. The accuracy for {\it 
interest} was approximately 81\% and {\it line} was disambiguated with 
slightly less than 80\% accuracy.  The lesson taken from these results was 
that an ensemble should consist of classifiers that represent 
as differently sized windows of context as possible; this  reduces the 
impact of redundant errors made by classifiers that represent  very 
similarly sized windows of context.  The ultimate success of an  ensemble 
depends on the ability to select classifiers that make  complementary 
errors. This  is discussed in  the  context of  combining  
part--of--speech taggers  in  \cite{BrillH98}.  They  provide a measure  
for assessing the complementarity of errors between two taggers that  
could be adapted for use with larger ensembles such as the one discussed  
here, which has nine members.  

\subsection{Disambiguation by majority vote}

In this paper ensemble disambiguation is based on a simple majority vote 
of the nine member classifiers.  

An alternative strategy is to weight each vote by the
estimated joint probability found by the Naive Bayesian
classifier. However, a preliminary study found that the  accuracy of a
Naive Bayesian ensemble using a weighted vote was poor. For {\it
interest}, it resulted in accuracy of 83\% while for {\it line} it
was 82\%.  The simple  majority vote resulted in accuracy of
89\% for  {\it  interest} and  88\% for {\it line}. 

\section{Future Work}

A number of issues have arisen in the course of this work that merit
further investigation. 

The simplicity of the contextual representation can lead to large
numbers of parameters in the Naive Bayesian model when using wide
windows of context. Some combination of stop-lists and stemming could
reduce the numbers of parameters and thus improve the overall quality
of the parameter estimates made from the training data. 

In addition to simple co--occurrence features, the use of collocation
features seems promising. These are distinct from co--occurrences in
that they are words that occur in close proximity to the ambiguous
word and do so to a degree that is judged statistically significant. 

One limitation of the majority vote in this paper is that there is no
mechanism for dealing with outcomes where no sense gets a majority of
the votes. This did not arise in this study but will certainly occur
as Naive Bayesian ensembles are applied to larger sets of data.    

Finally, further experimentation with the size of the windows of
context seems warranted. The current formulation is based on a
combination of intuition and empirical study. An algorithm to
determine optimal windows sizes is currently under development. 

\section{Conclusions}

This paper shows that word sense disambiguation accuracy can be improved
by combining a number of simple classifiers into an ensemble.  A 
methodology for formulating an ensemble of Naive Bayesian classifiers
is presented, where each member classifier is based on co--occurrence
features extracted from a different sized window of context. 
This approach was evaluated using the widely studied nouns {\it line} and
{\it interest}, which are disambiguated with accuracy of  88\% and  89\%, 
which rivals the best previously published results. 

\section{Acknowledgments}

This work extends ideas that began in collaboration with Rebecca Bruce
and Janyce Wiebe. Claudia Leacock and Raymond Mooney provided
valuable assistance with the {\it line} data.  I am indebted to an
anonymous  reviewer who pointed out the importance of separate {\tt
test} and {\tt devtest} data sets. 

A preliminary version of this paper appears in \cite{Pedersen00a}.

\end{document}